\documentclass{ecai} 

\usepackage{latexsym}
\usepackage{amssymb}
\usepackage{amsmath}
\usepackage{amsthm}
\usepackage{booktabs}
\usepackage{enumitem}
\usepackage{graphicx}
\usepackage{color}
\usepackage{algorithm}
\usepackage{algorithmic}

\usepackage{multirow}
\usepackage{latexsym}
\usepackage{amssymb}
\usepackage{amsmath}
\usepackage{amsthm}
\usepackage{booktabs}
\usepackage{enumitem}
\usepackage{graphicx}
\usepackage{color}
\usepackage{utfsym}
\usepackage{makecell}
\usepackage{float}
\usepackage{stfloats}
\usepackage{url}
\usepackage[hidelinks]{hyperref}

\newcommand{\BibTeX}{B\kern-.05em{\sc i\kern-.025em b}\kern-.08em\TeX}

\begin{document}


\begin{frontmatter}


\paperid{0847} 


\title{TADoc: Robust Time-Aware Document Image Dewarping}


\author[A,C]{\fnms{Fangmin}~\snm{Zhao}}
\author[A,C]{\fnms{Weichao}~\snm{Zeng}}
\author[A,C]{\fnms{Zhenhang}~\snm{Li}}
\author[A]{\fnms{Dongbao}~\snm{Yang}}
\author[B]{\fnms{Yu}~\snm{Zhou}\thanks{Corresponding Author. Email: yzhou@nankai.edu.cn.}}


\address[A]{Institute of Information Engineering, Chinese Academy of Sciences}
\address[B]{VCIP \& TMCC \& DISSec, College of Computer Science \& College of Cryptology and Cyber Science, Nankai University}
\address[C]{School of Cyber Security, University of Chinese Academy of Sciences}


\begin{abstract}
Flattening curved, wrinkled, and rotated document images captured by portable photographing devices, termed document image dewarping, has become an increasingly important task with the rise of digital economy and online working. Although many methods have been proposed recently, they often struggle to achieve satisfactory results when confronted with intricate document structures and higher degrees of deformation in real-world scenarios. Our main insight is that, unlike other document restoration tasks (e.g., deblurring), dewarping in real physical scenes is a progressive motion rather than a one-step transformation. Based on this, we have undertaken two key initiatives. Firstly, we reformulate this task, modeling it for the first time as a dynamic process that encompasses a series of intermediate states. Secondly, we design a lightweight framework called TADoc (\textbf{T}ime-\textbf{A}ware \textbf{Doc}ument Dewarping Network) to address the geometric distortion of document images.
In addition, due to the inadequacy of OCR metrics for document images containing sparse text, the comprehensiveness of evaluation is insufficient. To address this shortcoming, we propose a new metric -- DLS (\textbf{D}ocument \textbf{L}ayout \textbf{S}imilarity) -- to evaluate the effectiveness of document dewarping in downstream tasks. Extensive experiments and in-depth evaluations have been conducted and the results indicate that our model possesses strong robustness, achieving superiority on several benchmarks with different document types and degrees of distortion.
\end{abstract}

\end{frontmatter}


\section{Introduction}

Document images captured by mobile photography devices suffer from various distortions and background interference due to uncontrolled paper deformations and camera position. These drawbacks not only affect the users' reading experience but also impede downstream tasks (e.g., layout analysis \cite{li2025beyond}, optical character recognition \cite{zheng2024cdistnet,yang2025ipad,qiao2020seed,wang2022tpsnet,lyu2025arbitrary}, document question answering\cite{shen_ijcai}),  wherein expert models are mostly trained on scanned documents. 
Therefore, document image dewarping, aimed at eliminating geometric distortions, has received increasing attention.
Traditional dewarping methods focus mainly on measuring the 3D shape of degraded documents. They often rely on carefully designed and calibrated hardware, such as stereo cameras and structured light projectors \cite{meng2014active}, which suffer from a vast lack of practicality. Other methods leverage manually designed low-level features such as shading \cite{zhang2009unified} and contours, yet limited performance has been achieved.

\begin{figure}[t]
\begin{center}
\includegraphics[width=0.8\columnwidth]{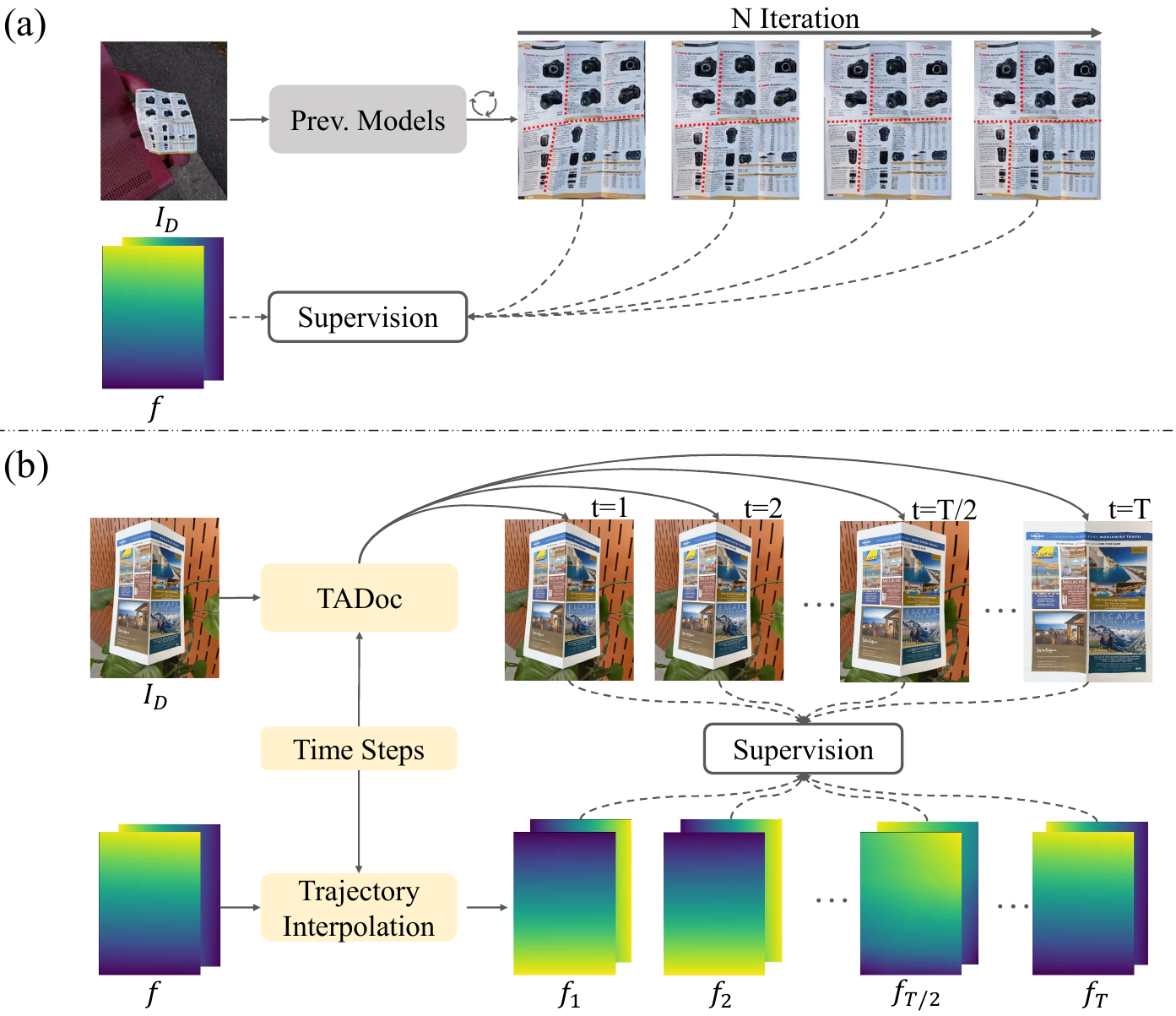}
\caption{Comparison of TADoc with previous progressive methods. For distorted image $I_{D}$: (a) Previous methods are trained with single mapping supervision $f$ with cascaded network. (b)  TADoc provides multi-step prediction and leverages dynamic distortion mapping $f_{t}$ for fine-grained supervision.}
\label{fig_intro}
\end{center}
\end{figure}

\begin{figure}[tb]
\begin{center}
\includegraphics[width=0.8\columnwidth]{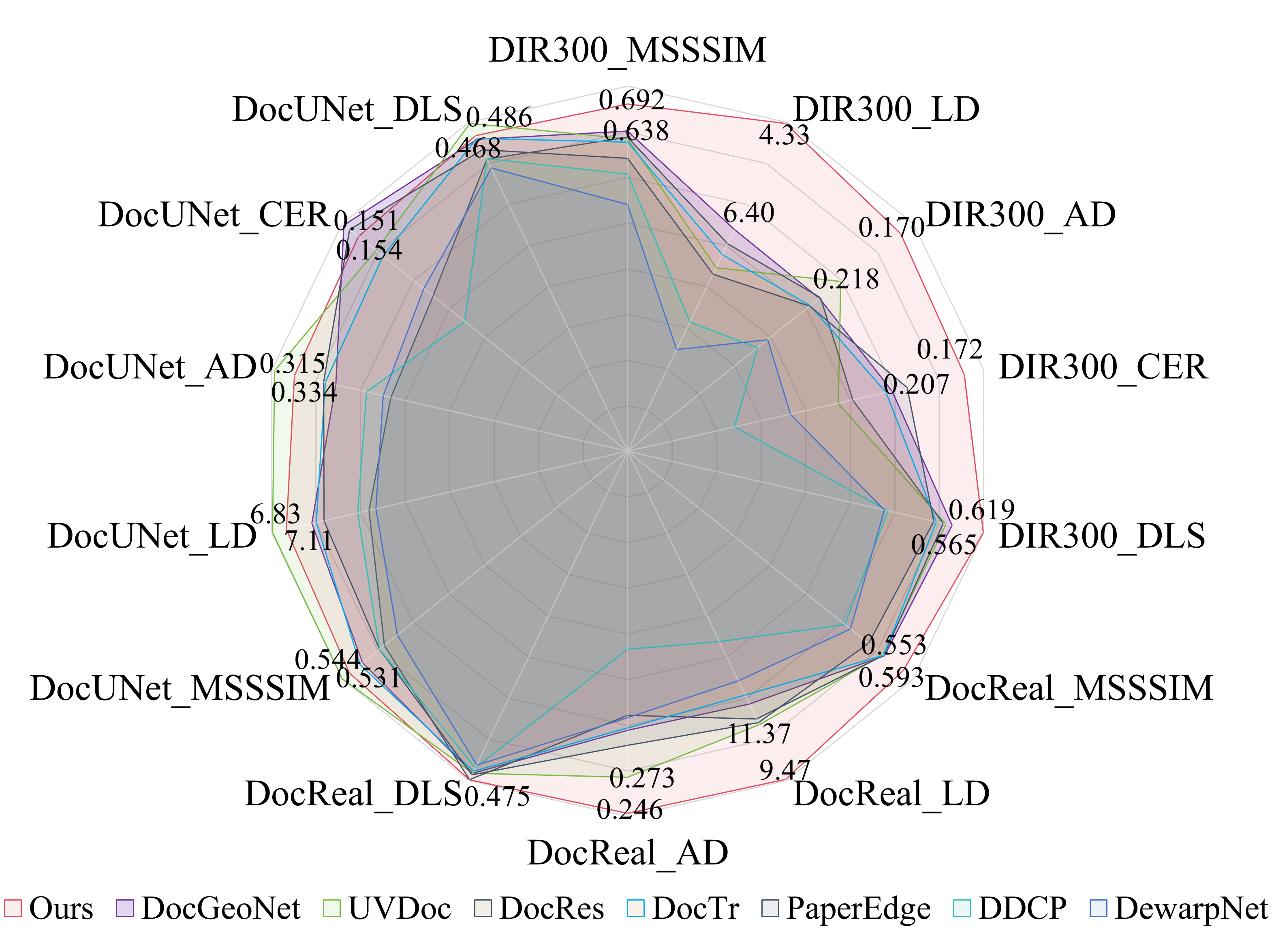}
\caption{The performance comparison on several benchmarks and metrics.}
\label{polar}
\end{center}
\end{figure}

Recently, methods based on deep learning have been proposed for document dewarping. These works \cite{ma2018docunet,feng2021doctr,xie2021document} typically model the task as a regression problem with a 2D vector field that represents the coordinate flow between distorted and flat images. Some methods have been proposed to flatten documents utilizing prior information such as 3D \cite{xu2022document}, text lines \cite{jiang2022revisiting} and layout \cite{li2023layout}. However, they require either additional complex annotations or time-consuming post-processing operations. 
Meanwhile, these models may suffer from degeneration when faced with a document with sparse textlines or unclear layout. As shown in Fig. \ref{polar}, previous methods fail to achieve satisfactory overall results on multiple datasets.

In this work, inspired by the scenario of flattening a distorted document in the real world, we reformulate the document dewarping task as a progressive rectification process. As shown in Fig. \ref{fig_intro} (b), we divide document dewarping into $T$ time steps, and as the time step gradually increases from $0$ to $T$, the document gradually recovers from the original distorted image to a completely flat image. 
This modeling has distinct advantages over previous progressive methods \cite{zhang2022marior}, as shown in Fig. \ref{fig_intro} (a). 
Firstly, previous methods use the results from the previous step as input, achieving a coarse-to-fine dewarping effect, which can lead to imbalanced learning difficulty across different stages and requires substantial time due to its sequential execution. In contrast, our method consistently takes the original distorted image as input. Given a time step, we compel the model to learn the corresponding degree of distortion, thereby enhancing its understanding of the distortion process and leading to superior performance. Additionally, our method supports parallel inference, with significantly higher processing speeds.

Based on multi-step temporal modeling, a lightweight model called TADoc (\textbf{T}ime-\textbf{A}ware \textbf{Doc}ument Dewarping Network) is proposed. Specifically, we introduce time embedding to extract time-aware distorted document features explicitly. Then, we decode the features and apply the predicted coordinate map to the original image to obtain different degrees of restored images. 
Fig. \ref{polar} displays the superiority of TADoc, with strong robustness to a wide range of document types and distortion levels.

In addition, previously OCR metrics for downstream evaluation fail to evaluate documents with sparse texts or no texts, leading to insufficient evaluation. To address the limitation, we propose a supplementary metric for evaluating the performance of document dewarping, termed Document Layout Similarity (DLS). Specifically, a pretrained document layout analysis (DLA) model is leveraged to process the dewarped document image and the corresponding scanned one, while DLS is defined as the Intersection-over-Union (IoU) of the analysis results. Compared with OCR, DLA is not only closely related to the users' reading experience, but can also be applied to all document images for comprehensive evaluation, thus better reflecting the quality of dewarping from a downstream perspective.

Overall, we have made three-fold contributions in this work for document image dewarping as follows:

\begin{itemize}
\item Inspired by the temporal progress of flattening in real-world scenarios, we reformulate document dewarping as a progressive process. To our knowledge, we are the first to introduce the concept of time steps into document image dewarping.
\item Based on the reformulation, we proposed a lightweight TADoc, which benefits from the temporal modeling to capture subtle distortion and achieves state-of-the-art performance on several commonly used benchmarks.
\item To alleviate the shortcoming of OCR assessment, we propose a new metric -- Document Layout Similarity (DLS), provides a comprehensive evaluation of dewarping results in downstream tasks.
\end{itemize}

\section{Related Works}

\subsection{Traditional Document Dewarping}
Traditional dewarping methods mainly focus on 3D reconstruction and 2D prior. Multi view images or additional hardwares are often used to assist in 3D reconstruction. Zhang et al. \cite{zhang2009unified} use a range/depth tensor to measure the 3D shape of the deformed papers. Meng et al. \cite{meng2014active} obtain the document curl using a platform with two structured laser beams. Tsoi and Brown \cite{tsoi2007multi} synthesize the flat image by combining the boundary of images from different perspectives. You et al. \cite{you2017multiview} model the crease information on multiple images and reconstruct the 3D shape. For 2D image processing, low-level features such as text lines are often emphasized. 
Mischke and Luther \cite{mischke2005document} model the detected text lines as polynomial curves, and  Lavialle et al. \cite{lavialle2001active} choose to model them as cubic B-splines.

\subsection{Deep Learning-Based Document Dewarping}
With the development of deep learning, many document dewarping methods \cite{ma2018docunet,feng2021doctr,kumari2024readable,shu2025visual,li2024first,zeng2024textctrl} based on deep learning have been proposed.
DocUNet \cite{ma2018docunet} and DocTr \cite{feng2021doctr} use CNN and Transformer architectures, respectively, to predict the global map. DewarpNet \cite{das2019dewarpnet} and CREASE \cite{markovitz2020can} first predicts the 3D coordinates, and then generates a dense backward map. 
DDCP \cite{xie2021document}, DocReal \cite{yu2024docreal} and UVDoc \cite{verhoeven2023uvdoc} proposes to predict sparse control point coordinates. Methods such as DocGeoNet \cite{feng2022geometric} and La-DocFlatten \cite{li2023layout} utilize prior information like edges, text lines, and layout to improve the dewarping effects. Marior \cite{zhang2022marior} proposes to dewarp images from coarse to fine. Tang et al. \cite{tang2024efficient} proposed a dual-stream architecture that addresses both geometric distortion and uneven illumination simultaneously, while Kumari et al. \cite{kumari2024readable} introduced the concept of transfer learning, applying geometric correction from natural scenes to the document domain. Zhang et al. \cite{zhang2024document} train image registration models to obtain real images with annotations, thereby improving dewarping effects from a data perspective. DocMamba \cite{han2025docmamba} achieves robust dewarping effect based on an selective state space sequence model.

\subsection{Document Layout Analysis}
DLA aims to recognize document layout and position.  Lebourgeois et al. \cite{lebourgeois1992fast} connect the characters into blocks, and then connect them into paragraphs. Journet et al. \cite{journet2005text} distinguish text and graphic regions by extracting directional features within smaller square areas. Diem et al. \cite{diem2011text} use clustering to classify document regions. Garz et al. \cite{garz2010detecting} use SIFT features to detect layouts. The deep learning approaches model DLA as a detection \cite{li2022dit} or segmentation task \cite{biswas2022docsegtr} using CNN or transformer frameworks, benefiting from the development of Faster-RCNN \cite{girshick2015fast}, Mask-RCNN \cite{he2017mask}, and others.

\section{Proposed Method}

\begin{figure*}[ht]
\begin{center}
\includegraphics[width=0.9\textwidth]{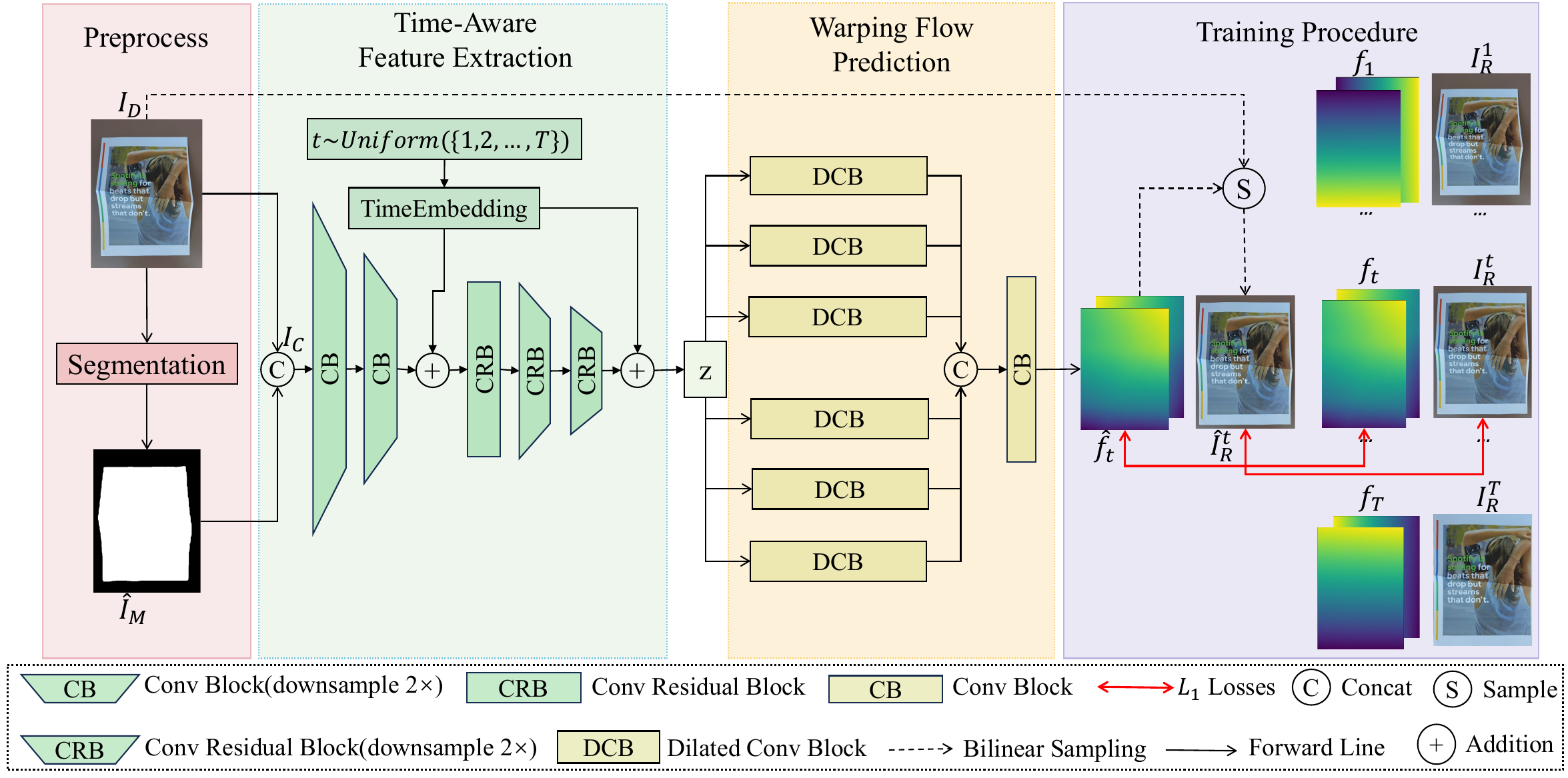}
\caption{An overview of the proposed TADoc. Three modules are contained, namely \textit{Preprocess}, \textit{Time-Aware Feature Extraction}, and \textit{Warping Flow Prediction}. Moreover, the training paradigm is also briefly illustrated.}
\label{overview}
\end{center}
\end{figure*}

Document dewarping in a progressive manner naturally coincides with the temporal progress of flattening in real-world scenarios. In this section, we will reformulate the problem, then detail the framework, TADoc, followed by the paradigm of training and inference.

\subsection{Problem Reformulation}

Given a distorted document image $I_D$, we obtain the restored image $I_R^t$ at each time step $t$ through the network by constructing the corresponding warping flow $\mathcal{F}_t$:
\begin{equation}
I_R^t = \mathcal{F}_t(I_D),
\end{equation} where $\mathcal{F}_t=\{ F^t(p)\mid p\in I_D \}$, $p$ is the pixel coordinate in the distorted document $I_D$ and the mapping $F^t(p)$ at each pixel coordinate $p$ from $I_D$ defines the position of that pixel at each time step $t$.

To further clarify, at time step $t$, $(x,y)$ and $(u_t,v_t)$ are the 2D coordinates of restored document $I_R^t$ and distorted document $I_D$, respectively. The relationship between two coordinates can be represented via a backward map $f_t$:
\begin{equation}
f_t(x,y)=(u_t,v_t),
\end{equation}

When the restoration process is divided into $T$ time steps, at time $t=0$, the document image $I_R^0$ is identical to $I_D$, with the corresponding backward map $f_0$ as:
\begin{equation}
f_0(x,y) = (x,y),
\end{equation}
wherein $f_0$ is a regular grid. 

At time $t=T$, the restored image $I_R^T$ is a completely flat image, with  $f_T$ written as:
\begin{equation}
f_T(x,y) = (u_T,v_T),
\end{equation}
wherein $f_T$ is equivalent to the backward map annotations.

For an intermediate time step $t$, the degree of image restoration varies from $f_0$ to $f_T$. For simplicity, we consider a linear interpolation to model the trajectory of transport between $f_0$ and $f_T$:
\begin{equation}
f_t = \frac{t}{T}f_T+(1-\frac{t}{T})f_0.
\label{ft}
\end{equation}

\subsection{Time-Aware Document Dewarping Network}

With the problem reformulated, a time-aware dewarping network is proposed, termed TADoc. It consists of Preprocessing Module, Time-Aware Feature Extraction (TAFE) Module and Warping Flow Prediction (WFP) Module, as depicted in Fig. \ref{overview}. We will delve into these modules in the following.

\textbf{Preprocess.} To eliminate the extra background texture from the distorted image $I_D \in \mathbb{R}^{H \times W \times 3}$, a segmentation network DeepLabv3+ \cite{chen2018encoder} is adopted to obtain the mask image $I_M \in \mathbb{R}^{H \times W}$. The auxiliary mask guides the model to focus on the content of informative areas.
Subsequently, we feed the original image $I_D$ along with the mask image $I_M$ into our proposed model to predict grid map. The preprocessing is fast and doesn't impact the inference speed.

\textbf{Time-Aware Feature Extraction.} The input of TAFE module is $I_C \in \mathbb{R}^{H \times W \times 4}$ and timestep $t$, wherein $I_C$ is obtained by concatenating $I_D$ with $I_M$. 
Compared with pixel-wise multiplication \cite{feng2021doctr}, the concatenation allows an adaptive integration of region information while alleviating the negative impact from imprecise masks.

In TAFE, $I_C$ is first fed into two convolutional layers to extract preliminary image information. Timestep $t$ is encoded as a time embedding and fused with preliminary image features. This is followed by multiple layers of residual blocks \cite{he2016deep} to get deeper document distortion features. Subsequently, this feature is added to the time embedding to enhance TADoc's ability to control the degree of distortion to obtain the time-aware distortion features $z \in \mathbb{R}^{\frac{H}{16} \times \frac{W}{16} \times c}$. 

\textbf{Warping Flow Prediction.} The WFP module aims to use the time-aware features $z$ obtained above to predict the warping flow. Inspired by DDCP \cite{xie2021document}, we choose to predict the backward map of sparse grid points, which makes our model lightweight without the need for upsampling operations. Specifically, $z$ go through six convolutional blocks with different dilation rates \cite{yu2015multi} respectively to enrich the network's insight ability into multi-scale and improve robustness. Subsequently, these features are aggregated and sent to a single-layer convolutional layer to obtain the warping flow $f_t \in \mathbb{R}^{\frac{H}{16} \times \frac{W}{16} \times 2}$ , which can be seen as the positions of $\frac{H}{16} \times \frac{W}{16}$ reference points in $I_D$. Finally, $f_t$ is bilinearly interpolated to the size of $I_D$ to obtain the restored document image $I_R^t$ at time $t$.

\textbf{Training Loss.} During the training stage, TADoc is optimized with the following objective, 
\begin{equation}
\mathcal{L} = \alpha||f_t-\hat{f_t}||+\beta\mathcal{L}_r.
\end{equation}
where $f_t$ and $\hat{f_t}$ are the annotated flow and the predicted counterparts. $\mathcal{L}_r$ is $L_1$ losses on the ground truth flat images and the dewarped images obtained through the predicted grid backward map $\Hat{f}_t$. $\alpha$ and $\beta$ are leveraged to balance the influence of the individual loss terms.

\begin{algorithm}[tb]
\caption{Training}
\label{training}
\textbf{Input:} Dataset $\mathcal{S}$\\
\textbf{Initiate:} Model $\mathcal{M}_{\theta}$
\begin{algorithmic}[1] 
\REPEAT
\STATE $t \sim \text{Uniform}(\{1,2, \ldots,T\})$
\STATE $I_D, f_T, I_R^t \sim \mathcal{S}$
\STATE $f_t = \frac{t}{T}f_T+(1-\frac{t}{T})f_0$
\STATE $\Hat{f}_t \leftarrow \mathcal{M}_{\theta}(I_D, t)$
\STATE $\Hat{f}_T = \frac{T}{t}\Hat{f}_t+(1-\frac{T}{t})f_0$
\STATE $\Hat{I}_R^t \leftarrow Sampling(I_D, \Hat{f}_T)$
\STATE Take gradient descent step on $\nabla L$
\UNTIL{converged}
\end{algorithmic}
\end{algorithm}

\subsection{Paradigm of Training and Inference}

Now we discuss about the training and inference processes.

Algorithm \ref{training} displays the complete training procedure. For each sample in the training set, we obtain distorted image $I_D$, ground truth backward map $f_T$, and ground truth flattened image $I_R$, and then randomly select a time step $t$ from $1$ to $T$ with equal probability to calculate the corresponding supervised signal $f_t$. Subsequently, we train the model until convergence, enabling it to have a more comprehensive understanding of the dewarping process.

During the inference phase, as shown in Algorithm \ref{inference}, we try two inference methods: direct method and average method. For direct method, we can simply input the original image $I_D$ and the final time step $t=T$ into the network to predict the backward map  $f_T^{direct}$. For average method, the predicted information from all $T$ time steps is adopted. Specifically, for every time step $t$, the trained TADoc predicts the backward map $\Hat{f}_t$, and the corresponding final map $\Hat{f}_T^t$ can be deduced following Eq. \ref{ft}:
\begin{equation}
\Hat{f}_T^t=\frac{T}{t}\Hat{f}_t+(1-\frac{T}{t})f_0,
\end{equation}

The result $f_T^{avg}$ of the average method is the average of the final flow obtained for all time steps:
\begin{equation}
f_T^{avg} = \frac{1}{T}\sum_{t=1}^{T}\Hat{f}_T^t.
\end{equation}
Finally, the flattened image $I_R^{direct}$ and $I_R^{avg}$  can be obtained by sampling from $I_D$ through $f_T^{direct}$ and $f_T^{avg}.$

\section{Experiments}

\subsection{Dataset}

TADoc is trained on the Doc3D and UVDoc datasets and evaluated on the DIR300, DocReal and DocUNet benchmarks. A brief introduction of each dataset is provided.

 \textbf{Training Data.} The Doc3D dataset contains 100K synthetic distorted document images with rich annotations, including 3D maps, surface normals, albedo maps, depth maps, and backward map. The UVDoc dataset contains 20K pseudo-photorealistic document images with 4,032 geometric deformations. Annotations involve sparse backward maps, sparse 3D grids and masks.

\begin{algorithm}[tb]
\caption{Parallel Inference}
\label{inference}
\textbf{Input:} Image $I_D$\\
\textbf{Parameter:} Trained Model $\mathcal{M}_{\theta}$
\begin{algorithmic}[1] 
\FORALL{$t=1,2,...,T$ in parallel}
    \STATE $\Hat{f}_t \leftarrow \mathcal{M}_{\theta}(I_D, t)$
    \STATE $\Hat{f}_T^t = \frac{T}{t}\Hat{f}_t+(1-\frac{T}{t})f_0$
\ENDFOR
\STATE $\Hat{f}_T^{sum} \leftarrow \sum_{t=1}^{T} \Hat{f}_T^t$
\STATE $f_T^{direct} = \Hat{f}_T^T$
\STATE $f_T^{avg} = \frac{1}{T}\Hat{f}_T^{sum}$
\STATE $I_R^{direct} \leftarrow Sampling(I_D, f_T^{direct})$
\STATE $I_R^{avg} \leftarrow Sampling(I_D, f_T^{avg})$
\end{algorithmic}
\end{algorithm}

 \textbf{Evaluation Data.} The DIR300 benchmark contains 300 real-world document photos captured by different mobile devices with various distortion types.
 The DocReal benchmark contains 200 Chinese document images, covering a wide range of real-world scenarios involving tables, contracts, books and so on. The DocUNet benchmark contains 130 images from 65 different documents covering receipts, letters, magazines and music sheets.

\subsection{Evaluation Metrics}

To ensure a comprehensive assessment of dewarping results, image similarity metrics, OCR metrics and the proposed DLS are adopted for quantitative evaluation in the experiments.

\begin{table*}[ht]
    \begin{center}
    \small
    \caption{Quantitative comparison of dewarping performance on the DIR300 benchmark. ``$\uparrow$'' and ``$\downarrow$'' signifies higher
and lower better respectively. The \textbf{best} results are represented in bold, the \underline{second best} results are represented in underline, the \underline{\underline{third best}} results are represented in double underline, and ``-'' indicates that the dewarped results are inaccessible due to lack of open-source availability. The same applies to the tables below.}
\vspace{+10pt}
    \begin{tabular}{cccccccccc}
        \toprule
        Method & Venue & MS-SSIM $\uparrow$ & LD $\downarrow$ & AD $\downarrow$ & CER $\downarrow$ &  ED $\downarrow$ & DLS $\uparrow$ & Parameter $\downarrow$ & FPS $\uparrow$\\
        \midrule
        DewarpNet & \textit{ICCV'19} & 0.492 & 13.94 & 0.331 & 0.356 & 1059.6 & 0.447 & 86.9M & 0.43\\
        DocTr & \textit{MM'21} & 0.616 & 7.21 & 0.254 & 0.224 & 699.6 & 0.537 & 26.9M & 0.64\\
        DDCP & \textit{ICDAR'21} & 0.552 & 10.95 & 0.357 & 0.541 & 2085.0 & 0.453 & 13.3M & 1.82\\
        PaperEdge & \textit{SIGGRAPH'22} & 0.584 & 8.00 & 0.255 & 0.533 & 508.7 & 0.324 & 36.6M & 0.50\\
        DocGeoNet & \textit{ECCV'22} & \underline{\underline{0.638}} & 6.40 & 0.242 & 0.565 & 665.0 & 0.375 & 24.8M & 0.59\\
        UVDoc & \textit{SIGGRAPH'23} & 0.622 & 7.73 & 0.218 & 0.275 & 675.4 & 0.554 & 8.0M & 0.87\\
        LA-DocFlatten & \textit{TOG'23} & 0.652 & \underline{\underline{5.70}} & 0.195 & 0.189 & 511.1 & \underline{\underline{0.589}} & - & -\\
        DocRes & \textit{CVPR'24} & 0.626 & 6.83 & 0.241 & 0.257 & 763.2 & 0.549 & 15.2M & 0.44\\
        DocTLNet & \textit{IJDAR'24} & \underline{0.658} & 5.75 & - & \underline{0.177} & \underline{482.6} &  - & - & -\\
        DocMamba & \textit{ICMM'25} & 0.632 & 6.57 & \underline{\underline{0.194}} & - & - & - & - & - \\
        \midrule
        Ours (direct) &   & \textbf{0.692} & \underline{4.39} & \underline{0.175} & \underline{\underline{0.180}} & \underline{\underline{499.0}} & \textbf{0.621} & 7.9M & 0.78\\
        Ours (average) &   & \textbf{0.692} & \textbf{4.33} & \textbf{0.170} & \textbf{0.172} & \textbf{475.0} &  \underline{0.619} & 7.9M & 0.78\\
        \bottomrule
    \end{tabular}
     \label{dir300}
\end{center}
\end{table*}

\begin{figure}[tb]
\begin{center}
\includegraphics[width=0.9\columnwidth]{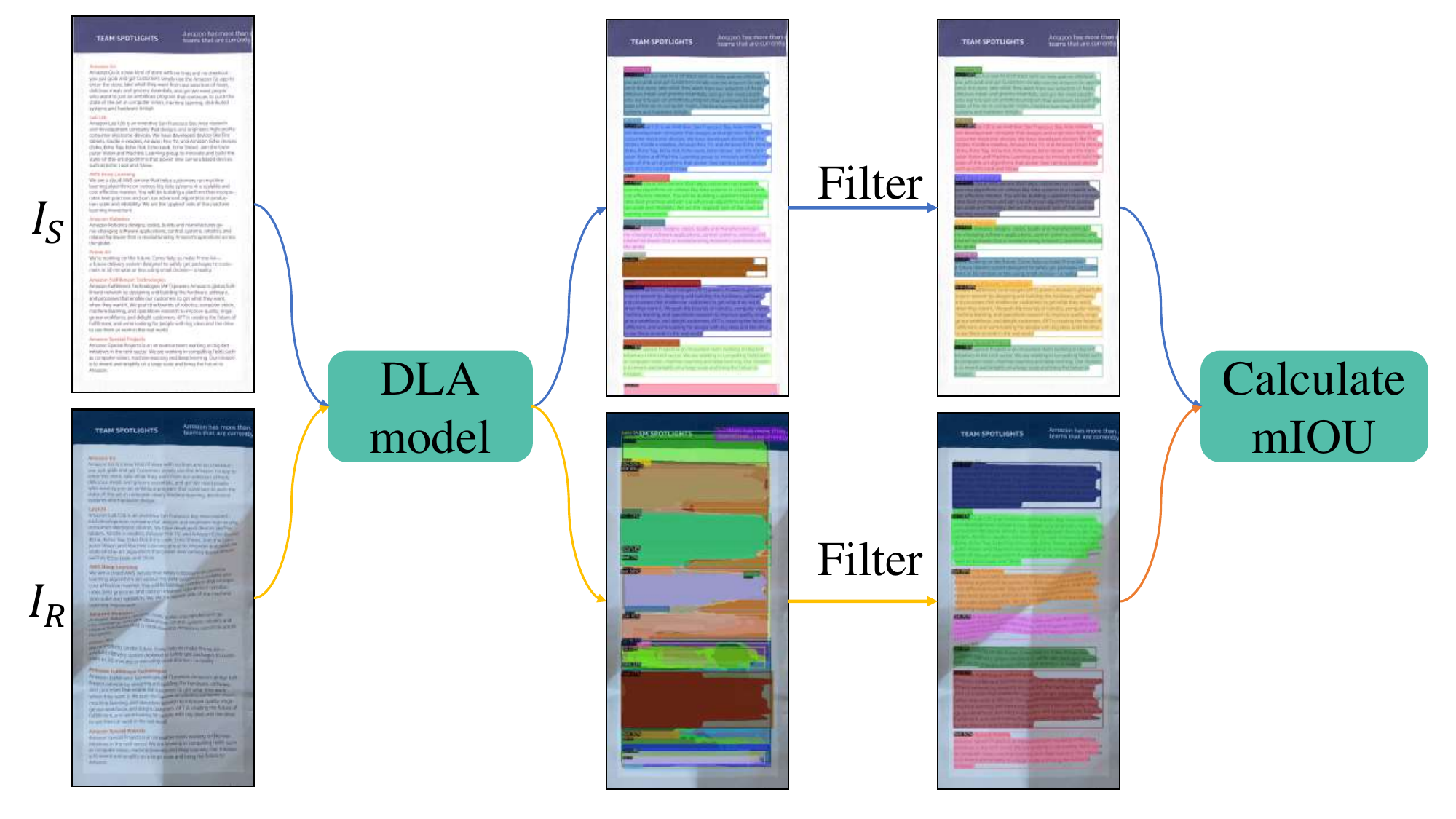} %
\caption{The pipeline of calculating DLS metric.}
\label{layout_iou}
\end{center}
\end{figure}

\textbf{Image Similarity Metrics.} MS-SSIM builds a Gaussian pyramid of two images and is defined as the weighted sum of SSIM across multiple resolutions. We follow the parameters in DocUNet \cite{ma2018docunet}. Local Distortion (LD) utilizes SIFT flow for dense image registration between two images, measuring the average deformation of each pixel. Aligned Distortion (AD) enhances MS-SSIM through unified translation and scaling operations while alleviating LD errors using image gradients as weighting factors.

\begin{figure}[tb]
\begin{center}
\includegraphics[width=0.9\columnwidth]{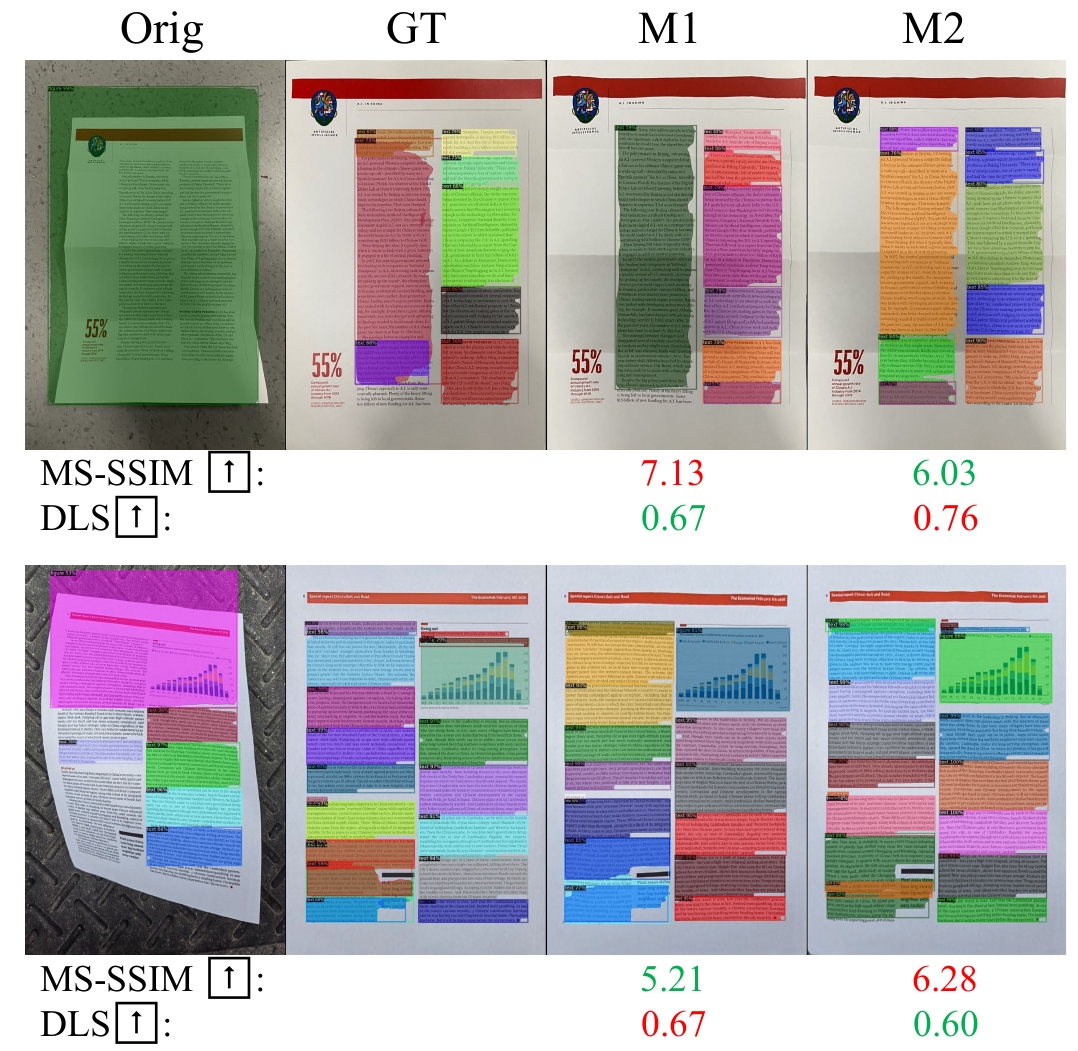}
\caption{Comparison of DLS and MS-SSIM result.}
\label{dls-ssim}
\end{center}
\end{figure}

\textbf{OCR Metrics.} Edit Distance (ED) is defined as the minimal number of substitutions (s), insertions(i), and deletions (d) to transform a string into the reference one. And Character Error Rate (CER) is defined as $CER = (s+i+d)/N$, where $N$ refers to the total number of characters in the reference text. 90 pairs of images in DIR300, 60 pairs of images in DocUNet and all images in DocReal are generally selected for OCR evaluation, consistent with previous experiments. The selected images contain a great amount of text, contributing to stable result of CER. We perform the evaluation with Tesseract v5.0.1 and PyTesseract v0.3.9.

\textbf{Document Layout Similarity.} 
Although OCR metrics are important indicators for evaluating the performance of dewarping results in downstream tasks, it cannot process documents with sparse texts or no texts (e.g., music sheets, illustrations). As mentioned above, only a subset of images from the benchmarks are qualified for evaluation, cannot comprehensively reflect the dewarping effects of all document types. Therefore, we turn to another downstream task with looser constraints, DLA. 
Based on DLA, we propose the DLS. As shown in Fig. \ref{layout_iou}, we leverage a pretrained model DiT to process scanned document $I_S$ and dewarped document $I_R$ pair, obtaining corresponding layout segmentation regions and categories.
Layout areas with a low confidence level are filtered for stability and credibility. Subsequently, for each corresponding layout area pair, the IOU is calculated. DLS represents the average value of IOU across all layout areas, which introduces a novel perspective to evaluate dewarping methods. In addition, while both DLS and MS-SSIM measure the similarity between two images, DLS places greater emphasis on high-level semantic similarity, whereas MS-SSIM focuses more on low-level similarity. Specifically, for identical layout regions in two images, if the DLA model classifies them into different categories, the DLS score will be zero. Overall, DLS imposes stricter requirements on semantic similarity.
As shown in Fig. \ref{dls-ssim}, for the dewarping result of two method M1 and M2, they cannot always perform optimally on both MS-SSIM and DLS. Although the result of M1 attains higher MS-SSIM, the result of M2 demonstrates better dewarping effects, particularly in text lines, and also achieves higher DLS score. Therefore, we believe that DLS serves as a supplementary metric and the introduction of DLS is both reasonable and necessary.

\subsection{Implementation Details}

We implement our model in PyTorch and train it on a NVIDIA GeForce RTX 4090. A batch size of 96 is used with the AdamW optimizer. The initial learning rate is set as $10^{-4}$ and weight decay is set of $10^{-4}$. 
The hyperparameters $\alpha$ and $\beta$ of the loss function are both set to 1. During the evaluation, we test FPS of different methods, all on a NVIDIA GeForce RTX 4090.

\subsection{Comparison with Prior Methods}

\begin{table*}[ht]
    \begin{center}
    \setlength{\tabcolsep}{1mm}
    \caption{Quantitative comparison of dewarping performance on the DocReal and DocUNet benchmark.}
    \small
    \vspace{+10pt}
    \begin{tabular}{cc|cccccc|cccccc}
        \toprule
        \multicolumn{2}{c}{ } & \multicolumn{6}{c}{DocReal} & \multicolumn{6}{c}{DocUNet}\\
        \hline
        Method & Venue & MS-SSIM $\uparrow$ & LD $\downarrow$ & AD $\downarrow$ & CER $\downarrow$ & ED $\downarrow$ & DLS $\uparrow$ & MS-SSIM $\uparrow$ & LD $\downarrow$ & AD $\downarrow$ & CER $\downarrow$ & ED $\downarrow$ & DLS $\uparrow$\\
        \hline
        DewarpNet & \textit{ICCV'19} & 0.480 & 13.54 & 0.334 & 0.781 & 561.2 & 0.454 & 0.439 & 9.63 & 0.455 & 0.210 & 525.5 & 0.420\\
        DocTr & \textit{MM'21} & 0.551 & 12.74 & 0.322 & 0.801 & 519.2 & 0.463 & 0.510 & 7.79 & 0.368 & 0.175 & 464.8 & 0.464\\
        DDCP & \textit{ICDAR'21} & 0.468 & 16.32 & 0.450 & 0.961 & 721.2 & 0.458 & 0.473 & 8.98 & 0.426 & 0.263 & 745.4 & 0.434\\
        PaperEdge & \textit{SIGGRAPH'22} & 0.519 & 11.45 & 0.303 & 0.733 & 507.3 & \underline{\underline{0.468}} & 0.473 & 7.99 & 0.366 & 0.154 & \underline{375.6} & 0.447\\
        Marior & \textit{MM'22} & - & - & - & - & - & - & 0.478 & 7.27 & 0.403 & 0.214 & 593.8 & 0.421\\
        DocGeoNet & \textit{ECCV'22} & 0.553 & 12.29 & 0.319 & \underline{\underline{0.733}} & 535.1 & 0.466 & 0.505 & 7.68 & 0.381 & \underline{0.151} & \underline{\underline{379.0}} & 0.463\\
        DocReal & \textit{WACV'23} & \underline{\underline{0.556}} & \underline{\underline{9.83}} & \textbf{0.238} & 0.758 & \textbf{466.7} & \underline{0.473} & 0.498 & 7.83 & \underline{0.311} & \textbf{0.144} & \textbf{360.3} & 0.463\\
        LA-DocFlatten & \textit{TOG'23} & - & - & - & - & - & - & 0.526 & \textbf{6.72} & \textbf{0.300} & \underline{\underline{0.153}} & 391.9 &  \underline{0.476}\\
        UVDoc & \textit{SIGGRAPH'23} & 0.550 & 11.37 & 0.273 & \underline{0.732} & 503.5 & 0.466 & \textbf{0.544} & \underline{6.83} & \underline{\underline{0.315}} & 0.170 & 461.1 & \textbf{0.486}\\
        DocRes & \textit{CVPR'24} & 0.551 & 11.61 & 0.337 & 0.742 & 550.8 & \textbf{0.475} & 0.464 & 9.37 & 0.470 & 0.220 & 598.8 & 0.433\\
        Tang et al. & \textit{ICASSP'24} & - & - & - & - & - & - & 0.490 & 7.96 & - & - & - & -\\
        DocMamba & \textit{ICMM'25} & - & - & - & - & - & - & 0.529 & \underline{\underline{7.07}} & 0.338 & - & - & - \\
        \hline
        Ours (direct) &   & \underline{0.591} & \underline{9.51} & \underline{\underline{0.247}} & 0.769 & \underline{\underline{494.1}} & \underline{0.473} & \underline{0.531} & 7.25 & 0.335 & \underline{\underline{0.153}} & 423.4 & 0.463\\
        Ours (average) &   & \textbf{0.593} & \textbf{9.47} & \underline{0.246} & \textbf{0.717} & \underline{493.3} & \textbf{0.475} & \underline{\underline{0.530}} & 7.11 & 0.334 & 0.159 & 420.5 & \underline{\underline{0.468}}\\
        \bottomrule
    \end{tabular}
    \label{docreal_docunet}
    \end{center}
\end{table*}

\begin{figure*}[tb]
\begin{center}
\includegraphics[width=0.9\textwidth]{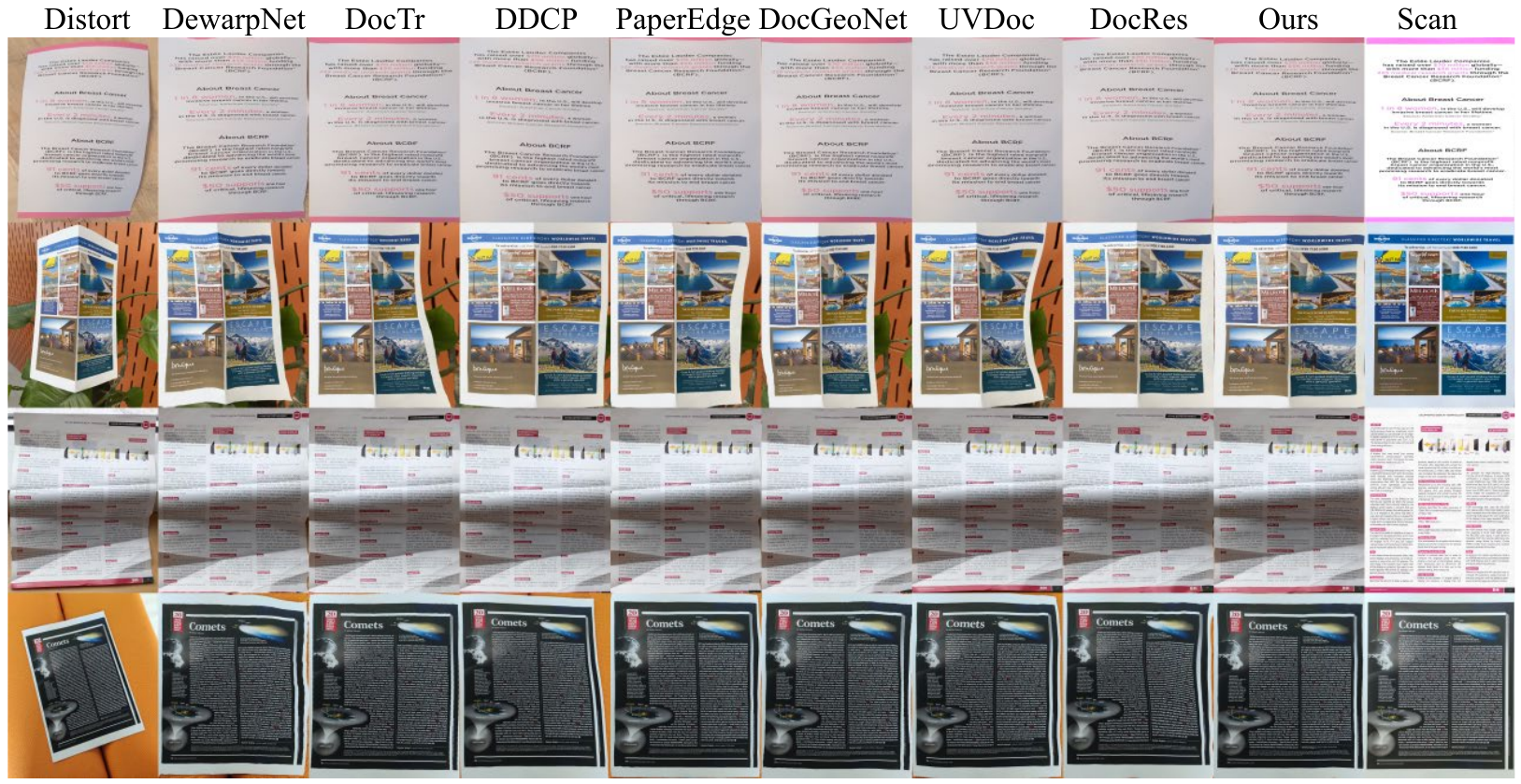}
\caption{Qualitative comparison with deep learning-based methods on distorted images from several benchmarks. Columns 2-9 represent the dewarping results of each method.}
\label{quan}
\end{center}
\end{figure*}

We quantitatively and qualitatively compare TADoc with deep learning based document dewarping approaches that have been open-sourced or provide available official results on the DIR300, DocReal and DocUNet benchmarks. Notably, we report the results of our model with both direct method and average method at a total time step of $20$. 

\textbf{Quantitative evaluation.} 
As shown in Tab. \ref{dir300} and Tab. \ref{docreal_docunet}, TADoc demonstrate superior overall performance. Specifically, on the DIR300, TADoc not only achieves the best results across all evaluation metrics but also significantly outperforms previous SOTA methods. Notably, the LD and AD metrics show substantial improvements, with reductions of 1.33 and 0.025, respectively. This also demonstrates that our dewarping results not only enhance the users' reading experience but also improve the performance of document images in downstream analysis tasks, including OCR and DLA. On the DocReal, TADoc surpasses previous SOTA on all metrics except AD and ED, where it ranks second, demonstrating robust results. On the DocUNet, we rank second in MS-SSIM, rank third in ED and DLS. In general, benefiting from the temporal reformulation and modeling, TADoc exhibits state-of-the-art or comparable performance on all benchmarks with various document types and deformation degrees.

\begin{figure*}[tb]
\begin{center}
\includegraphics[width=0.8\textwidth]{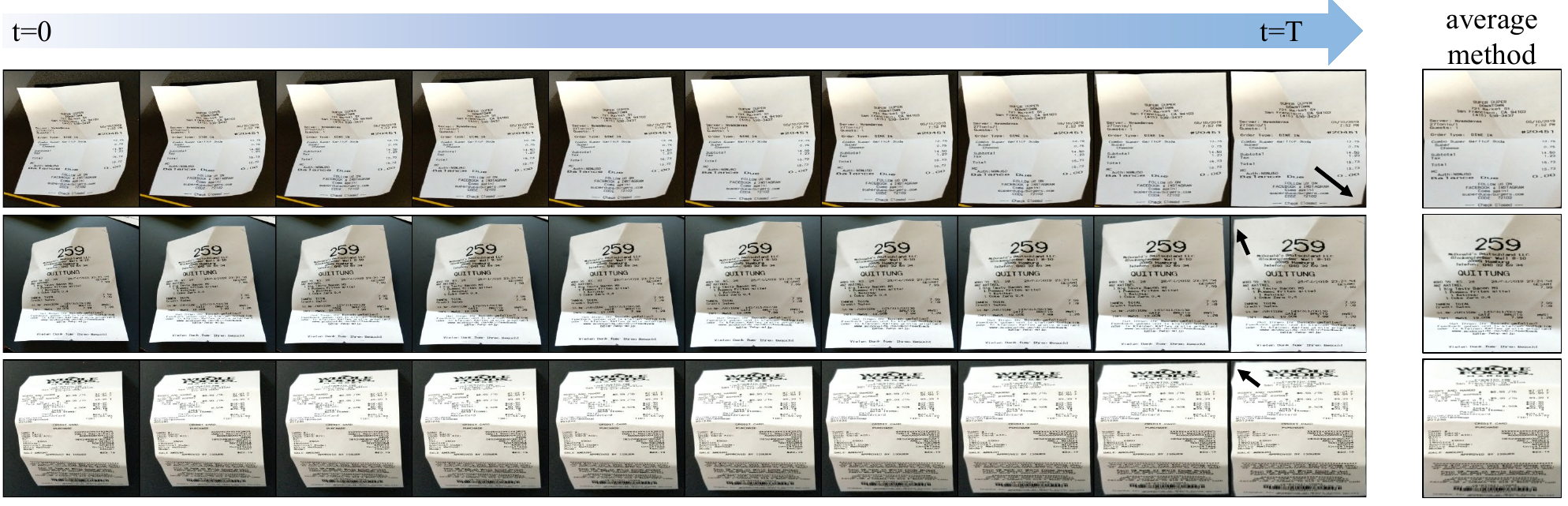}
\caption{visualization of TADoc's restored result at various times as well as the final images obtained through the average method on real-world documents.}
\label{realcase}
\end{center}
\end{figure*}

\begin{figure}[tb]
\begin{center}
\includegraphics[width=\columnwidth]{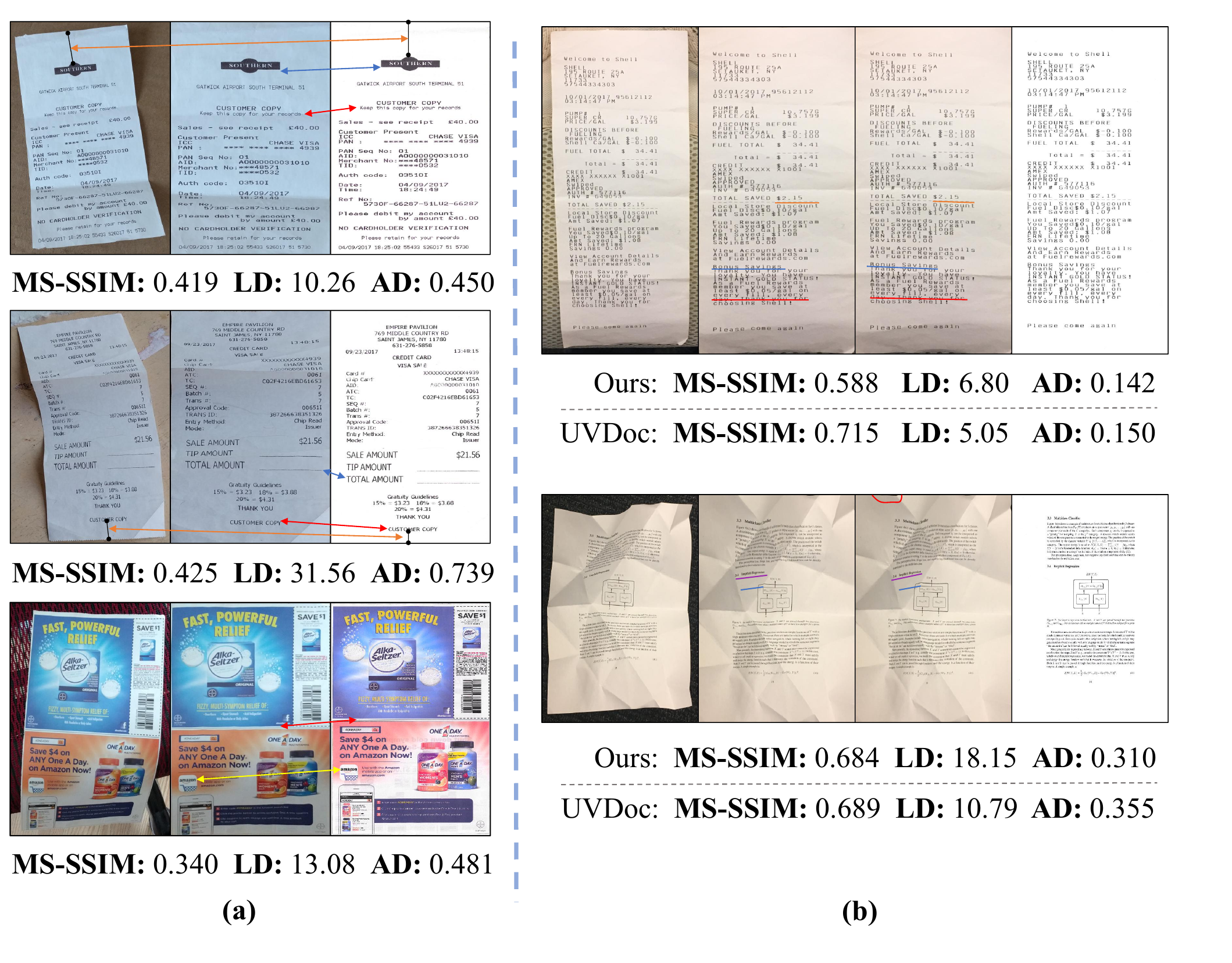}
\caption{Bad case analysis on DocUNet. (a) From left to right: degraded image, TADoc dewarping image, and scanned image. (b) From left to right: degraded image, TADoc dewarping image, UVDoc dewarping image, and scanned image.}
\label{docunet_analyse}
\end{center}
\end{figure}

Notably, no method has achieved a complete advantage on DocUNet. Apart from the OCR metric, which could not be evaluated for all images, we investigate the reasons related to image similarity. As observed by DocGeoNet \cite{feng2022geometric}, the reference images in DocUNet may suffer from imperfect alignment due to scanning error, which can lead to significant performance degradation, particularly affecting the LD metric. As shown in Fig. \ref{docunet_analyse} (a), TADoc performs well in restoring document edges and text lines but its quantitative metrics are notably poor. This misalignment issue also affects the comparison of various methods. As shown in Fig. \ref{docunet_analyse} (b), TADoc performs better in aligning text lines and edges on some images but exhibits poorer quantitative metrics. Despite this, we have achieved competitive results on DocUNet. Overall, the comprehensive performance across three datasets demonstrates that TADoc delivers superior and robust results.

In addition, Tab. \ref{dir300} also displays the model parameters and average inference speed of each method on DIR300. Due to our simple model architecture, streamlined pipeline, and method design for predicting sparse grid point warping flow, TADoc is able to achieve excellent results with minimal parameters. Even though the average method requires 20 steps of inference, parallel inference addresses this issue effectively.

\begin{table}[ht]
    \begin{center}
    \small
    \caption{Ablation study of timestep $T$ and loss function on DIR300.}
    \vspace{+10pt}
    \begin{tabular}{cccccccc}
        \toprule
        $T$ & $\beta$ & MS-SSIM $\uparrow$ & LD $\downarrow$ & AD $\downarrow$ & DLS $\uparrow$\\
        \midrule
        $1$ & 1 & 0.673 & 4.95 & 0.193 & 0.590\\
        $5$ & 1 &0.676 & 4.81 & 0.188 & 0.599\\
        $10$ & 1 & 0.682 & 4.64 & 0.186 & 0.603\\
        $20$ & 1 & 0.692 & 4.33 & 0.170 & 0.619\\
        $20$ & 0.5 & 0.690 & 4.35 & 0.175 & 0.622\\
        20 & 0.1 & 0.684 & \textbf{4.29} & 0.182 & 0.615\\
        20 & 0 & 0.691 & 4.41 & 0.172 & 0.613\\
        $30$ & 1 & \textbf{0.694} & 4.30 & \textbf{0.166} & \textbf{0.627}\\
        \bottomrule
    \end{tabular}
    \label{ablation}
    \end{center}
\end{table}

\textbf{Qualitative evaluation.} We further compare the visual results of different dewarping methods side by side in Fig. \ref{quan}. The document types of the selected images include flyers, posters, and magazines, while the deformation types involve rotation, bending, and folding. 
As depicted, previous methods lack robustness on various document types as well as distortion types and struggle with subtle deformation at the edge or corners, while our method works well on all these images. Our results not only excel in image edge restoration, but also achieve high-quality preservation of text paragraph structures and horizontal alignment of text lines. These improvements are particularly significant as our method does not rely on any prior knowledge of text lines or document layouts, unlike previous methods that incorporated such priors.

As shown in the Fig. \ref{realcase}, we demonstrate TADoc's capability to restore various degrees of distortion at different times, illustrating the intermediate steps of our dewarping process. The last column presents the results of real-world images using the average method. It is evident that the average method provides more robust dewarping, particularly excelling in the restoration of text lines and edges. This highlights the superiority of our modeling, which allows for a more precise perception of the distortion extent, thereby achieving more robust dewarping results. Further visualization of DLS and more qualitative results on the benchmarks, and more analysis can be found in the appendix.


\textbf{Ablation Studies.} In ablation, we delve into the effects of total time steps $T$ and loss function design, as well as the benefits our modeling method on DocTr. Specifically, different numbers of total time steps $T=1,5,10,20,30$ are selected for model training respectively, and results are evaluated using the average method with corresponding $T$ on the DIR300 benchmark. It is worth noting that the case of $T=1$ degenerates TADoc to previous one-step modeling methods, which directly predicts the final warping flow of flat images from distorted images. 
As shown in Tab. \ref{ablation}, increasing the total number of time steps $T$ can improve the effectiveness of model dewarping, yet the improvement becomes marginal beyond 20. Since the performance at $T=20$ is already satisfactory, and considering the trade-off between effectiveness and training resources, we report the results with $T=20$ for comparison. Next, we conduct an ablation study on the weight $\beta$ of the reconstruction loss $\mathcal{L}_r$. Training TADoc with the $L_1$ losses on the backward map simply leads to a worse performance due to the lack of detailed guidance from images.
Next, we conduct training on DocTr using our modeling method, with other settings kept consistent with the baseline, in order to validate the effectiveness of our method. As shown in the Tab. \ref{doctr}, our approach achieves significant improvement over the baseline on the DIR300 dataset, demonstrating the superiority of our modeling strategy.

\begin{table}[ht]
    \begin{center}
    \caption{Ablation study of our modeling for DocTr on DIR300.}
    \vspace{+10pt}
    \scriptsize
    \begin{tabular}{cccc}
        \toprule
        Method & MS-SSIM $\uparrow$ & LD $\downarrow$ & AD $\downarrow$\\
        \midrule
        DocTr \textit{(baseline)} & 6.160 & 7.214 & 0.254\\
        DocTr \textit{(Ours)} & 6.657 ($\uparrow$ 8\%) & 5.509 ($\downarrow$ 29\%) & 0.203 ($\downarrow$ 20\%) \\
        \bottomrule
    \end{tabular}
        \label{doctr}
    \end{center}
\end{table}

\subsection{Limitations}
Despite the superior performance, limitations exist in our method. Firstly, we focus on the geometric rectification of the document, while the illumination rectification still needs to be addressed. 
Secondly, as a preliminary exploration, the dewarping process is formulated with linear interpolation out of intuition. We will explore a variation of interpolation in modeling and corresponding performance in the future.

\section{Conclusion}

In this work, we have rethought and modeled the document dewarping task to make it more in line with the laws of real physical scenes. To our knowledge, we are the first to introduce the concept of time steps in document dewarping tasks. 
Based on this, we propose a lightweight network termed TADoc with time embedding to predict intermediate processes of dewarping. 
Our method achieves state-of-the-art results on multiple benchmarks, with strong robustness confronted with various document types and deformation degrees. 
Besides, we introduce a novel metric DLS for comprehensively measuring geometric dewarping performance.

\begin{ack}
 This work is supported by the National Natural Science Foundation of China (Grant NO 62376266 and 62406318).
\end{ack}



\bibliography{mybibfile}

\end{document}